\documentclass{article} 
\usepackage{nips12submit_e,times}

\usepackage{amsmath}
\usepackage{amsfonts}
\usepackage{latexsym}
\usepackage[ruled,vlined]{algorithm2e}

\usepackage{graphicx}

\usepackage[T1]{fontenc} 

\usepackage{tabu}

\title{Prediction of breast cancer recurrence using Classification Restricted Boltzmann Machine with \textit{Dropping}}
\author{
Jakub M.~Tomczak \\
Wroc\l aw University of Technology \\
Wroc\l aw, Poland \\
\texttt{jakub.tomczak@pwr.wroc.pl} \\
}

\nipsfinalcopy 

\graphicspath{ {images/} }

\begin{document}

\maketitle

\begin{abstract}
In this paper, we apply Classification Restricted Boltzmann Machine (ClassRBM) to the problem of predicting breast cancer recurrence. According to the Polish National Cancer Registry, in 2010 only, the breast cancer caused almost $25\%$ of all diagnosed cases of cancer in Poland. We propose how to use ClassRBM for predicting breast cancer return and discovering relevant inputs (symptoms) in illness reappearance. Next, we outline a general probabilistic framework for learning Boltzmann machines with masks, which we refer to as \textit{Dropping}. The fashion of generating masks leads to different learning methods, i.e., \textit{DropOut}, \textit{DropConnect}. We propose a new method called \textit{DropPart} which is a generalization of \textit{DropConnect}. In \textit{DropPart} the Beta distribution instead of Bernoulli distribution in \textit{DropConnect} is used. At the end, we carry out an experiment using real-life dataset consisting of 949 cases, provided by the Institute of Oncology Ljubljana.
\end{abstract}

\section{Introduction}

Machine learning algorithms has been successfully applied to many complex problems, especially in medical domain \cite{K:01}, e.g., in diabetes treatment \cite{TG:13}, in predicting breast cancer recurrence \cite{SBKZKG:10}. In this paper, we would like to focus on the second problem since according to Polish National Cancer Registry, in 2010 only, $25\%$ of all cases of cancer were associated with the breast cancer \cite{NCR}. Current diagnostics techniques are unable to successfully predict breast cancer return, therefore, there is a constant need to develop new predictive models. According to the author's knowledge, Classification Restricted Boltzmann Machine (ClassRBM) has been not yet applied to the prediction of the breast cancer recurrence. Moreover, the ClassRBM can be used to determine relevant symptoms of the illness reappearance. These two aspects constitute an original contribution to the field of machine learning and modeling of biomedical phenomena.

In this paper, only a preliminary study on learning ClassRBM with \textit{Dropping} is presented. However, the paper aims at making the following contribution:
\begin{itemize}
\item We propose a general probabilistic framework for learning ClassRBM with masks, which we call \textit{Dropping}. The fashion of generating masks leads to different learning methods, namely, \textit{DropOut} \cite{HSKSS:12,S:13} and \textit{DropConnect} \cite{WZZLF:13}.
\item We propose a new learning method which is a generalization of \textit{DropConnect}. The connections are partially removed during learning, i.e., an activation of each connection is drawn with Beta distribution (with $a, b \leq 1$). We call this method \textit{DropPart}.
\item We use ClassRBM to discover relevant inputs, or symptoms in the medical domain.
\item We carry out an experiment using real-life dataset consisting of 949 cases, provided by the Institute of Oncology Ljubljana \cite{SBKZKG:10}.
\end{itemize}

The paper is structured as follows. In Section \ref{sect:model} Classification Restricted Boltzmann Machine is outlined. In Section \ref{sect:prediction} the conditional distribution for prediction is given. In Section \ref{sect:prediction} the conditional distribution for discovering relevant inputs is described. In Section \ref{sect:learning} learning with \textit{Dropping} is outlined. At the end of the paper, an experiment that examines the predictive and explanatory capabilities of ClassRBM in in the problem of breast cancer recurrence is carried out. Obtained results within the experiment are discussed and conclusions are drawn.

\section{Classification Restricted Boltzmann Machine}

\subsection{The model}
\label{sect:model}

Restricted Boltzmann Machine (RBM) is a two-layer undirected graphical model where the first layer consists of visible input variables $\mathbf{x} \in \{0,1\}^{D}$, and the second layer consists of hidden variables (units) $\mathbf{h} \in \{0,1\}^{M}$. We allow only the inter-layer connections, i.e., there are no connections within layers. Moreover, we add a third layer that represents observable output variable $y \in \{1, 2, \ldots, K\}$. Further, we use the 1-to-$K$ coding scheme which results in representing output as a binary vector of length $K$ denoted by $\mathbf{y}$, such that if the output (or class) is $k$, then all elements are zero except element $y_{k}$ which takes the value $1$.

A RBM with $M$ hidden units is a parametric model of the joint distribution of visible and hidden variables, that takes the form:
\begin{equation}\label{eq:RBM}
p( \mathbf{x}, \mathbf{y}, \mathbf{h} | \boldsymbol\theta ) = \frac{1}{Z(\boldsymbol\theta)} e^{-E(\mathbf{x}, \mathbf{y}, \mathbf{h}|\boldsymbol\theta)}
\end{equation}

with parameters $\boldsymbol\theta = \{ \mathbf{b}, \mathbf{c}, \mathbf{d}, \mathbf{W}^{1}, \mathbf{W}^{2} \}$, and where:

\begin{equation}\label{eq:energy}
E(\mathbf{x}, \mathbf{y}, \mathbf{h} | \boldsymbol\theta) = -\mathbf{b}^{\top}\mathbf{x} - \mathbf{c}^{\top}\mathbf{h} - \mathbf{d}^{\top}\mathbf{y} - \mathbf{x}^{\top} \mathbf{W}^{1} \mathbf{h} - \mathbf{h}^{\top} \mathbf{W}^{2} \mathbf{y}
\end{equation}

is an energy function, and

\begin{equation}\label{eq:partition_function}
Z(\boldsymbol\theta) = \sum_{\mathbf{x}, \mathbf{y}, \mathbf{h}} e^{ -E(\mathbf{x}, \mathbf{y}, \mathbf{h} | \boldsymbol\theta) }
\end{equation}

is a partition function.

This model is called Classification Restricted Boltzmann Machine (ClassRBM)\footnote{The ClassRBM was first proposed in \cite{LB:08}.} and is argued to be used as a stand-alone non-linear classifier \cite{LMPB:12}. The main advantage of using RBM as a classifier is that it remains all generative advantages of RBM and additionally it allows to calculate distribution $p(y|\mathbf{x})$ straightforwardly. It can be shown that the following expressions hold true for ClassRBM \cite{LB:08, LMPB:12}:\footnote{Further in the paper, sometimes we omit explicit conditioning on parameters $\boldsymbol\theta$.}

\begin{align}
p(\mathbf{x} | \mathbf{h}) &= \prod_{i} p(x_{i}|\mathbf{h}) \label{eq:x_h} \\
p(x_{i}=1 | \mathbf{h}) &= \mathrm{sigm}( b_{i} + \mathbf{W}_{i \cdot}^{1} \mathbf{h} ) \label{eq:xd_h} \\
p(y | \mathbf{h}) &= \frac{ e^{d_{y} + (\mathbf{W}_{\cdot y}^{2})^{\top} \mathbf{h} } }{ \sum_{\overline{y}} e^{d_{\overline{y}} + (\mathbf{W}_{\cdot \overline{y}}^{2})^{\top} \mathbf{h} } } \label{eq:inputs} \\
p(\mathbf{h} | y, \mathbf{x}) &= \prod_{j} p(h_{j}|\mathbf{h}) \label{eq:h_yx} \\
p(h_{j} = 1 | y, \mathbf{x}) &= \prod_{j} \mathrm{sigm}( c_{j} + W_{jy}^{2} + (\mathbf{W}_{\cdot j}^{1})^{\top} \mathbf{x} ) \label{eq:hm_yx}
\end{align}

where $\mathrm{sigm}(\cdot)$ is the logistic sigmoid function, $\mathbf{W}_{i \cdot}^{\ell}$ is $i^{\text{th}}$ row of weights matrix $\mathbf{W}^{\ell}$, $\mathbf{W}_{\cdot j}^{\ell}$ is $j^{\text{th}}$ column of weights matrix $\mathbf{W}^{\ell}$, $W_{ij}^{\ell}$ is the element of weights matrix $\mathbf{W}^{\ell}$.

\subsection{Prediction}
\label{sect:prediction}

It is possible to exactly compute the distribution $p(y|\mathbf{x})$ which can be further used to choose the most probable class label. This conditional distribution takes the following form \cite{LB:08, LMPB:12}:
\begin{equation}\label{eq:y_x}
p(y|\mathbf{x}) = \frac{ e^{d_{y}} \prod_{j} \big{(} 1 + (e^{ c_{j} + (\mathbf{W}_{\cdot j}^{1})^{\top} \mathbf{x} })\ e^{ W_{jy}^{2} } \big{)}  }{ \sum_{\overline{y}} e^{d_{\overline{y}}} \prod_{j} \big{(} 1 + (e^{ c_{j} + (\mathbf{W}_{\cdot j}^{1})^{\top} \mathbf{x} })\ e^{ W_{j\overline{y}}^{2} } \big{)} }.
\end{equation}

Pre-computing the terms $c_{j} + (\mathbf{W}_{\cdot j}^{1})^{\top} \mathbf{x}$ allows to reduce the time needed for computing the conditional distribution to $O(MD + MK)$ \cite{LB:08, LMPB:12}.

\subsection{Discovering relevant inputs}
\label{sect:discovery}

We can also use the ClassRBM to determine relevancy of inputs by calculating conditional probabilities $p(x_{i} = 1 | \mathbf{x}_{\setminus i}, y)$ \cite{LMPB:12}. In the medical context this conditional probability expresses the probability of occurring $i^{\text{th}}$ input for given other inputs (symptoms) and class label. For example, if $y=1$ denotes breast cancer recurrence, we can quantitatively determine which inputs are important (relevant) during prediction. Further, we make an assumption that for $i^{\text{th}}$ input all other inputs are inactive, i.e, $\mathbf{x}_{\setminus i} = \mathbf{0}$. Hence, we get the following expression for the conditional probability:
\begin{align}\label{eq:x_y}
p(x_{i} = 1|\mathbf{x}_{\setminus i} = \mathbf{0}, y) &= \frac{ e^{b_{i}} \prod_{j} \big{(} 1 + e^{ c_{i} + (\mathbf{W}_{\cdot i}^{1})^{\top} \mathbf{x} + W_{jy}^{2} } \big{)} }{ \sum_{x_{i} = \{0,1\}} e^{b_{i}} \prod_{j} \big{(} 1 + e^{ c_{j} + (\mathbf{W}_{\cdot j}^{1})^{\top} \mathbf{x} + W_{j\overline{y}}^{2} } \big{)} } \nonumber\\
 &=  \frac{ e^{b_{i}} \prod_{j} \big{(} 1 + e^{ c_{j} + (W_{i j}^{1}) + W_{jy}^{2} } \big{)} }{ e^{b_{d}} \prod_{j} \big{(} 1 + e^{ c_{j} + (W_{i j}^{1}) + W_{jy}^{2} } \big{)} + \prod_{j} \big{(} 1 + e^{ c_{j} + W_{jy}^{2} } \big{)}\big{)} }
\end{align}

Calculating the conditional probabilities for all inputs using Equation \ref{eq:x_y} leads to discovering relevant inputs. For example, we may choose a threshold, e.g., equal $0.5$, and inputs with probabilities which are larger than the threshold are supposed to be important in the prediction.

\subsection{Learning with \textit{Dropping}}
\label{sect:learning}

Typically, the parameters $\boldsymbol\theta$ in ClassRBM are learned from data using the likelihood function:
\begin{equation}\label{eq:likelihood_typical}
p(\mathbf{x}, \mathbf{y}|\boldsymbol\theta) = \frac{1}{Z(\boldsymbol\theta)} \sum_{\mathbf{h}}e^{-E(\mathbf{x}, \mathbf{y}, \mathbf{h}|\boldsymbol\theta)}.
\end{equation}

There are different inductive principles for learning ClassRBM \cite{MSCF:10}, e.g., Approximate Maximum Likelihood, Maximum Pseudo-Likelihood, Ratio Matching. However, the widely-used method is Contrastive Divergence \cite{H:02, H:10}, which is further used in this paper.

Recently, new learning methods were introduced which perform a kind of regularization during learning RBM, namely, \textit{DropOut} \cite{HSKSS:12,S:13} and \textit{DropConnect} \cite{WZZLF:13}. In this paper, we would like to address this kind of learning and give a general approach called \textit{Dropping} which explains theoretical foundations of these two methods and gives rise to a new method which we refer to as \textit{DropPart}.

Let us introduce a \textit{mask}, $\mathcal{M} = \{ \mathbf{M}^{1}, \mathbf{M}^{2}, \mathbf{m} \}$, that determines which connections should be active during one learning iteration ($\mathbf{M}^{1}$ -- determines connections in weights $\mathbf{W}^{1}$, $\mathbf{M}^{2}$ -- determines connections in weights $\mathbf{W}^{1}$, and $\mathbf{m}$ -- connections in bias $\mathbf{c}$). Then, we get (symbol $\star$ denotes Hadamard product, called also element-wise product):
\begin{align}\label{eq:likelihood_mask}
p(\mathbf{x}, \mathbf{y}|\boldsymbol\theta) &= \sum_{\mathcal{M}} p(\mathbf{x}, \mathbf{y}, \mathcal{M}|\boldsymbol\theta) \nonumber \\
	&= \sum_{\mathcal{M}} p(\mathbf{x}, \mathbf{y}| \mathbf{b}, \mathbf{m} \star \mathbf{c}, \mathbf{d}, \mathbf{M}^{1} \star \mathbf{W}^{1}, \mathbf{M}^{2} \star \mathbf{W}^{2} )\ p(\mathcal{M} | \boldsymbol\theta ).
\end{align}

We may assume unconditional independence $\mathcal{M} \perp \boldsymbol\theta \mid \emptyset$ which yields
\begin{equation}\label{eq:likelihood_mask_independent}
p(\mathbf{x}, \mathbf{y}|\boldsymbol\theta) = \sum_{\mathcal{M}} p(\mathbf{x}, \mathbf{y}| \mathbf{b}, \mathbf{m} \star \mathbf{c}, \mathbf{d}, \mathbf{M}^{1} \star \mathbf{W}^{1}, \mathbf{M}^{2} \star \mathbf{W}^{2} )\ p(\mathcal{M} ).
\end{equation}

In general, we refer the approach of including mask in further inference to as \textit{Dropping}. The way the masks are generated yields different schemas:
\begin{itemize}
\item In \textit{Dropout} choosing a unit to be active is a Bernoulli random variable with probability $p$. Hence, if $i^{\text{th}}$ unit is active, then the $i^{\text{th}}$ row in $\mathbf{M}^{\ell}$ consists of ones ($\mathbf{M}^{\ell}_{i\cdot} = \mathbf{1}$), and zeros -- otherwise, for $\ell = 1, 2$. Similarly, if $i^{\text{th}}$ is drawn to be active, we set $\mathbf{m}_{i} = 1$. Typically, $p$ is set to $0.5$ \cite{HSKSS:12}.

\item In \textit{Dropconnect} choosing a connection to be active is a Bernoulli random variable, i.e., $\mathbf{M}^{\ell}_{ij} \sim \mathrm{Bernoulli}(p)$, for $\ell = 1, 2$, and $\mathbf{m}_{i} \sim \mathrm{Bernoulli}(p)$. The empirical results show that for $p=0.5$ the best performance was obtained \cite{WZZLF:13}.

\item We propose a new method called \textit{DropPart}. In \textit{DropPart} choosing a connection to be active is a Beta random variable, i.e., $\mathbf{M}^{\ell}_{ij} \sim \mathrm{Beta}(a,b)$, for $\ell = 1, 2$, and $\mathbf{m}_{i} \sim \mathrm{Beta}(a,b)$. The motivation for applying Beta distribution is twofold. First, for $a,b \leq 1$ one obtains a smooth version of \textit{DropConnect}. Second, such approach models spreading of bioelectric current in real neurons, i.e., neurons are active only partially (stronger or weaker).
\end{itemize}

During learning using Contrastive Divergence (or other gradient-based learning rule), it is important to calculate gradient of the log-likelihood function. Let us assume the likelihood function in form as in Equation \ref{eq:likelihood_mask_independent}. Then the logarithm of the likelihood function is intractable analytically, however, we can calculate its lower bound using the Jensen's inequality:
\begin{align}\label{eq:log_likelihood_mask_independent_lower_bound}
\log p(\mathbf{x}, \mathbf{y}|\boldsymbol\theta) &= \log \sum_{\mathcal{M}} p(\mathbf{x}, \mathbf{y}| \mathbf{b}, \mathbf{m} \star \mathbf{c}, \mathbf{d}, \mathbf{M}^{1} \star \mathbf{W}^{1}, \mathbf{M}^{2} \star \mathbf{W}^{2} )\ p(\mathcal{M} ) \nonumber \\
	&\geq \sum_{\mathcal{M}} \big{(} \log p(\mathbf{x}, \mathbf{y}| \mathbf{b}, \mathbf{m} \star \mathbf{c}, \mathbf{d}, \mathbf{M}^{1} \star \mathbf{W}^{1}, \mathbf{M}^{2} \star \mathbf{W}^{2} ) + \log p(\mathcal{M} ) \big{)}.
\end{align}

Notice that calculating the derivative of the lower bound of the log-likelihood in Equation \ref{eq:log_likelihood_mask_independent_lower_bound} wrt. $\mathbf{W}^{1}$, $\mathbf{W}^{2}$, $\mathbf{b}$, $\mathbf{c}$, and $\mathbf{d}$ is similar to calculating the derivative of the standard likelihood function in Equation \ref{eq:likelihood_typical} (because of the assumption of unconditional independence $\mathcal{M} \perp \boldsymbol\theta \mid \emptyset$ the term $\log p(\mathcal{M} )$ disappears) but with the summation over all possible masks $\mathcal{M}$. Moreover, the learning rules remain the same as in standard Contrastive Divergence (or other gradient-based learning procedures) but the mask is included, i.e., $\mathbf{m} \star \mathbf{c}$, $\mathbf{M}^{1} \star \mathbf{W}^{1}$, and $\mathbf{M}^{2} \star \mathbf{W}^{2}$.

During one learning iteration, in \textit{DropOut} and \textit{DropConnect} a crude approximation of the lower bound of the log-likelihood is used because the summation is replaced by one randomly drawn mask. Similarly, we use only one randomly drawn mask in \textit{DropPart}.

Since we have applied mask to learning, we should use mask in prediction. However, we omit this issue and leave it for further research. In this paper, we make predictions using the conditional distribution in Equation \ref{eq:y_x}. Only for \textit{DropOut} we apply the distribution in Equation \ref{eq:y_x} with weights $\mathbf{W}^{2}$ replaced by $\mathbf{W}^{2}/2$ as indicated in \cite{HSKSS:12}.


\section{Experiments}

\subsection{Preliminaries}

In the experiment we use the real-world medical dataset provided by the Institute of Oncology, Ljubljana \cite{SBKZKG:10}. The goal is to predict whether the patient will have a recurrence of a breast cancer within 10 years after surgery. Each patient is described by $15$ categorical features (details can be found in \cite{SBKZKG:10}).

In the experiment we used the \textit{classification accuracy} ($CA$) as the assessment metric. The $CA$ was applied because the authors of \cite{SBKZKG:10} have obtained results for two human doctors ($O1$ and $O2$ in Figure \ref{fig:accuracy}). The oncologists were asked to predict the class value for randomly chosen $100$ cases and then the $CA$ value was calculated \cite{SBKZKG:10}. The obtained quantities by oncologists are worst than those achieved by machine learning methods (see Figure \ref{fig:accuracy}) but this fact does not lead to a conclusion that the classifiers have significantly higher accuracy. However, it can give an insight in the usefulness of the application of machine learning methods in the medical domain.

We used the following classifiers in the experiment:
\begin{itemize}
\item ClassRBM with Contrastive Divergence learning (ClassRBM);
\item ClassRBM with Contrastive Divergence and DropOut learning (ClassRBM+DropOut);
\item ClassRBM with Contrastive Divergence and DropConnect learning (ClassRBM+DropConnect);
\item ClassRBM with Contrastive Divergence and DropPart learning (ClassRBM+DropPart);
\item Classification and Regression Tree (CART);
\item Naive Bayes classifier (Naive Bayes);
\item Bagging of $50$ CART (Bagging);
\item AdaBoost of CART (AdaBoost);
\item LogitBoost of CART (LogitBoost);
\item SVM with radial basis function (SVM);
\item Random Forest of $50$ CART (Random Forest).
\end{itemize}
ClassRBM and learning procedures were implemented in Matlab and for all other methods Matlab's implementations were used.

In order to have proper comparison of machine learning methods and human oncologists, we have used the original division of data into training set ($70\%$ of cases) and test set (remaining cases). All methods were learned and evaluated using this data division.

ClassRBM was learned using learning rate equal $0.01$ and $0.1$, momentum rate equal $0.5$, and $100 000$ number of iterations (no mini-batch technique was applied). Additionally, we used $\mathrm{Bern}(0.5)$ in \textit{DropOut} and \textit{DropConnect}, and three different sets of parameters in \textit{DropPart} with $\mathrm{Beta}(a,b)$, namely, $(a,b) \in \{(0.1, 0.1), (0.5, 0.5), (1, 1)\}$. All catagorical features were binarized resulting in $55$ binary inputs (see Appendix). The experiment was run $10$ times.

\subsection{Results and Discussion}

\begin{figure}[!htp]
\centering
\includegraphics[width=0.75\textwidth]{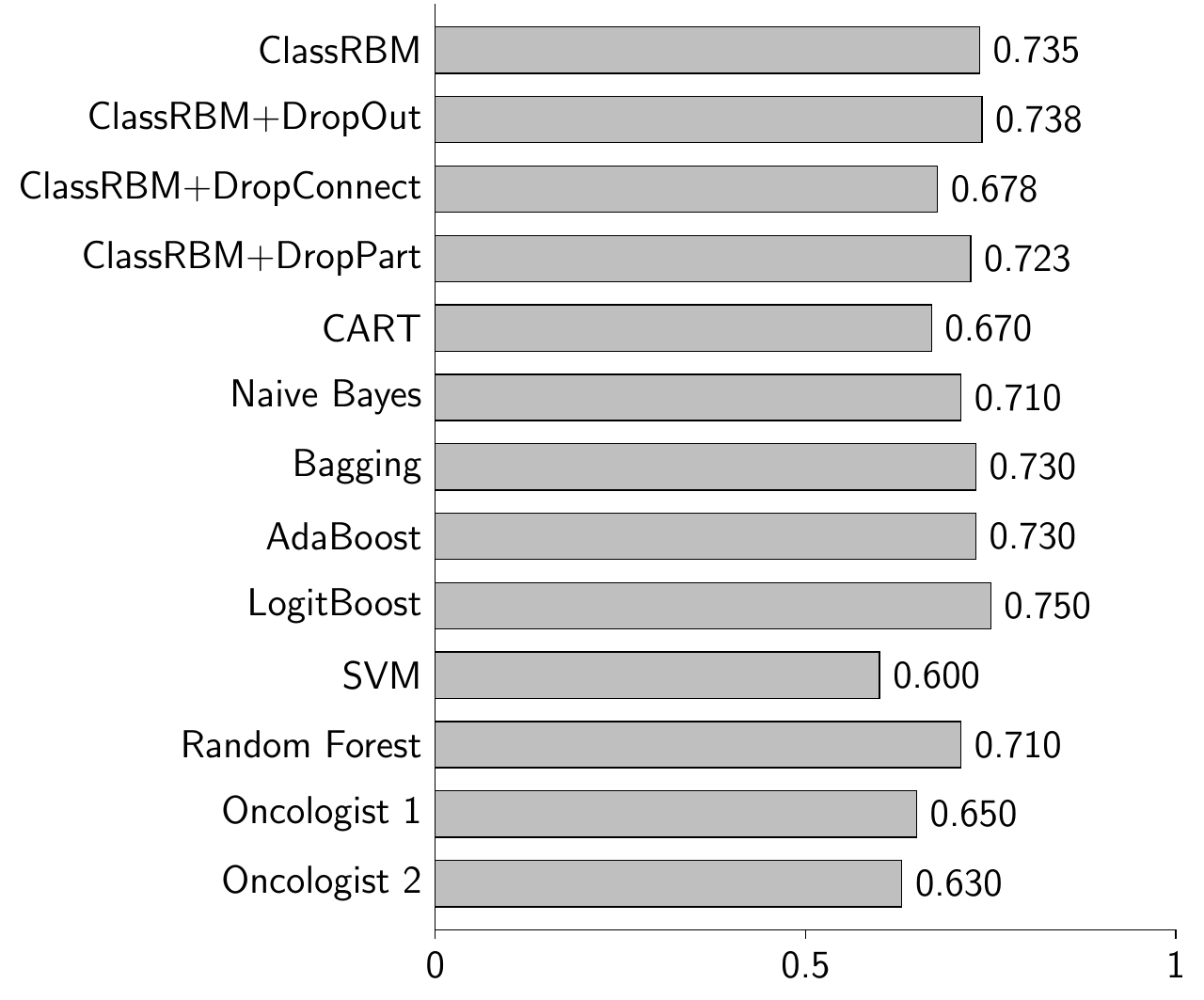}
\caption{Classification accuracy of considered models and two human oncologists. Note that the results for oncologists were obtained using part of test set (100 cases only) \cite{SBKZKG:10}.}
\label{fig:accuracy}
\end{figure}

The results for ClassRBM are given in Tables \ref{tab:accuracy1} and \ref{tab:accuracy2}. The best results for ClassRBM with different learning methods in comparison to other methods used in the experiment are given in Figure \ref{fig:accuracy}.

It can be noticed (see Figure \ref{fig:accuracy}) that ClassRBM performs comparably to ensemble classifiers (Bagging, AdaBoost and LogitBoost) and slightly better than Naive Bayes and Random Forest. It is remarkable because the ClassRBM obtains not only high classification accuracy but also provides relevant inputs discovery and assigns probabilities to class labels. These advantages of ClassRBM allows to state that it is high quality classifier basing on its quantitative (i.e. classification accuracy) and qualitative performance (i.e. generative capabilities).

It is worth to notice that \textit{DropOut} performed the best in comparison to \textit{DropConnect} and \textit{DropPart}. Additionally, comparing results for \textit{DropConnect} and \textit{DropPart} with $\mathrm{Beta}(0.1,0.1)$ (see Tables \ref{tab:accuracy1} and \ref{tab:accuracy2}) we can state that indeed \textit{DropPart} behaves very similar to \textit{DropConnect} as predicted. However, \textit{DropPart} performed better than \textit{DropConnect}. Nevertheless, we believe that application of an inference method which approximates the conditional probability in Equation \ref{eq:y_x} but with sum over masks will give much better results for \textit{DropConnect} and \textit{DropPart}. An approximated method for inference with \textit{DropConnect} is proposed in \cite{WZZLF:13}.

\begin{center}
\begin{table}[!htbp]
\caption{Results for ClassRBM with different learning methods and learning rate equal $0.01$. Mean values and standard deviations are given.}
\smallskip
\scriptsize{
\begin{tabu} to \linewidth {|c|cccccc|}
\hline
Number of		&  	ClassRBM	& ClassRBM 				& ClassRBM 				& ClassRBM					& ClassRBM 					& ClassRBM \\
hidden units	& 				& +DropOut 				& +DropConnect			& +DropPart 				& +DropPart  				& +DropPart \\
				& 				& $\mathrm{Bern}(0.5)$	& $\mathrm{Bern}(0.5)$	& $\mathrm{Beta}(0.1,0.1)$	& $\mathrm{Beta}(0.5,0.5)$	& $\mathrm{Beta}(1,1)$ \\
\hline
5  & $0.548 \pm 0.115$  & $0.679 \pm 0.037$  & $0.658 \pm 0.017$ & $0.671 \pm 0.026$ & $0.662 \pm 0.047$ & $0.620 \pm 0$ \\
10  & $0.685 \pm 0.114$ & $0.718 \pm 0.018$  & $0.666 \pm 0.025$ & $0.715 \pm 0.026$ & $0.719 \pm 0.031$ & $0.723 \pm 0.025$ \\
15  & $0.735 \pm 0.026$ & $0.738 \pm 0.015$  & $0.678 \pm 0.022$ & $0.702 \pm 0.035$ & $0.706 \pm 0.039$ & $0.720 \pm 0.027$ \\
20 & $0.714 \pm 0.025$  & $0.721 \pm 0.025$  & $0.677 \pm 0.013$ & $0.692 \pm 0.033$ & $0.717 \pm 0.028$ & $0.714 \pm 0.022$ \\
\hline
\end{tabu}
\scriptsize}
\label{tab:accuracy1}
\end{table}
\end{center}

\begin{center}
\begin{table}[!htbp]
\caption{Results for ClassRBM with different learning methods and learning rate equal $0.1$. Mean values and standard deviations are given.}
\smallskip
\scriptsize{
\begin{tabu} to \linewidth {|c|cccccc|}
\hline
Number of		&  	ClassRBM	& ClassRBM 				& ClassRBM 				& ClassRBM					& ClassRBM 					& ClassRBM \\
hidden units	& 				& +DropOut 				& +DropConnect			& +DropPart 				& +DropPart  				& +DropPart \\
				& 				& $\mathrm{Bern}(0.5)$	& $\mathrm{Bern}(0.5)$	& $\mathrm{Beta}(0.1,0.1)$	& $\mathrm{Beta}(0.5,0.5)$	& $\mathrm{Beta}(1,1)$ \\
\hline
5  & $0.714 \pm 0.044$ 	& $0.641 \pm 0.108$  & $0.553 \pm 0.119$ & $0.630 \pm 0.063$ & $0.696 \pm 0.034$ & $0.687 \pm 0.050$ \\
10  & $0.725 \pm 0.011$ & $0.694 \pm 0.014$  & $0.577 \pm 0.104$ & $0.649 \pm 0.022$ & $0.694 \pm 0.016$ & $0.701 \pm 0.016$ \\
15  & $0.704 \pm 0.028$ & $0.692 \pm 0.025$  & $0.606 \pm 0.080$ & $0.580 \pm 0.102$ & $0.687 \pm 0.019$ & $0.715 \pm 0.014$ \\
20 & $0.679 \pm 0.064$  & $0.679 \pm 0.063$  & $0.583 \pm 0.107$ & $0.643 \pm 0.114$ & $0.665 \pm 0.061$ & $0.704 \pm 0.021$ \\
\hline
\end{tabu}
\scriptsize}
\label{tab:accuracy2}
\end{table}
\end{center}

\begin{figure}[!htp]
\centering
\includegraphics[width=1\textwidth]{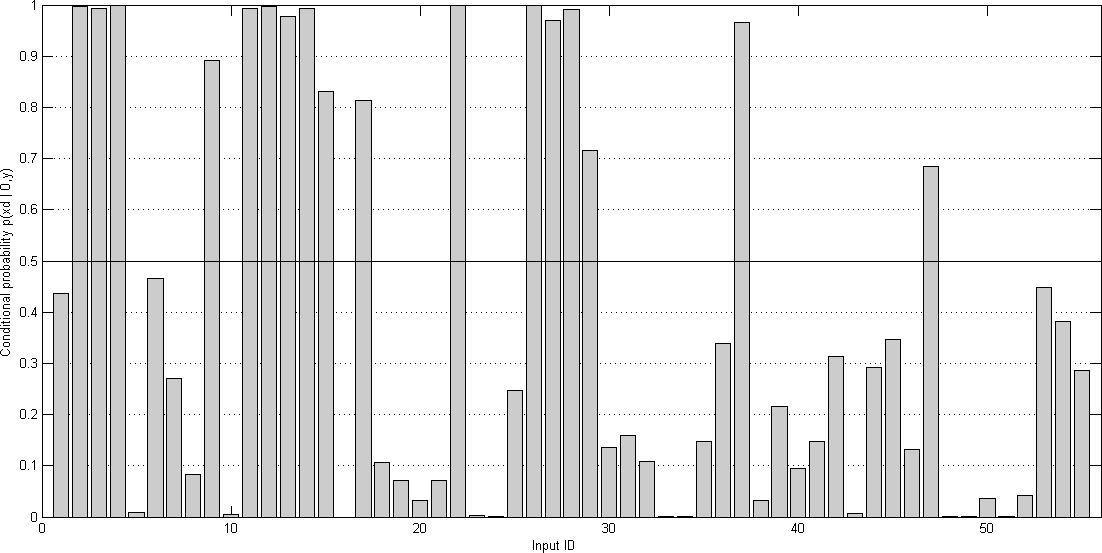}
\caption{Discovered relevant inputs for ClassRBM with \textit{DropPart}. The threshold is denoted as a solid horizontal line.}
\label{fig:relevantFeatures}
\end{figure}

At the end, we have applied the ClassRBM with \textit{DropPart} to discover relevant inputs (see Figure \ref{fig:relevantFeatures}). Because we obtained probabilities of all inputs given the information about the breast cancer return and assuming absence of other inputs. For example, it turned out that inputs Nos. $11$, $12$, $13$, i.e., histological type of tumor: ductual, lobular, and other, are important. Similarly, input No. $14$, i.e., level of progesterone receptors in tumor (in fmol per mg of protein) less than 10, is also relevant. On the other hand, it seems to be quite surprising that tumor stage over 50 mm (input No. $5$) does not matter in the breast cancer recurrence. However, medical doctors should determine now whether discovered features are interesting from medical or clinical point of view.

\section{Conclusions}

In this preliminary study we showed how to apply Classification Restricted Boltzmann Machine to the problem of predicting the breast cancer recurrence. We proposed the general probabilistic framework for learning with masks, called \textit{Dropping}. The fashion of generating masks leads to different learning methods, i.e, \textit{DropOut}, \textit{DropConnect}, and a new method called \textit{DropPart}. Our considerations are presented for ClassRBM but it can be straightforwardly applied to RBM, Deep Networks or Deep Boltzmann Machines. 

In this paper, we do not address several important issues, e.g., how to perform prediction with masks (e.g. an approximate inference using Gaussian approximation is proposed in \cite{WZZLF:13}), how to perform one learning iteration properly (see sum in Equation \ref{eq:log_likelihood_mask_independent_lower_bound}). Additionally, the performance of the methods can be improved by implementing some optimization tricks, e.g., mini-batch \cite{H:02}, centering trick \cite{MM:12}. Moreover, we have fixed the parameters values, however, we have an impression that in \textit{DropConnect} and \textit{DropPart} more learning iterations should be used in order to obtain more stable and thus better results. Last but not least, more thorough empirical studies are needed. We leave investigating all of these issues for further research.

\subsubsection*{Acknowledgments}
The research conducted by Jakub M. Tomczak has been partially co-financed by the European Union within the European Social Fund.

\bibliographystyle{abbrv}
\bibliography{onko_rbm}

\newpage

\section*{Appendix}
\scriptsize{
Description of all binarized inputs:
\begin{enumerate}
\item menopausal status \textit{false},
\item menopausal status \textit{true},
\item Tumor stage less than 20 mm,
\item Tumor stage between 20 and 50 mm,
\item Tumor stage over 50 mm,
\item Tumor grade good,
\item Tumor grade medium,
\item Tumor grade poor,
\item Tumor grade not applicable,
\item Tumor grade not determined,
\item Histological type of the tumor ductal,
\item Histological type of the tumor lobular,
\item Histological type of the tumor other,
\item Level of progesterone receptors in tumor (in fmol per mg of protein) less than 10,
\item Level of progesterone receptors in tumor (in fmol per mg of protein) more than 10,
\item Level of progesterone receptors in tumor (in fmol per mg of protein) unknown,
\item Invasiveness of the tumor no,
\item Invasiveness of the tumor invades the skin,
\item Invasiveness of the tumor invades the mamilla,
\item Invasiveness of the tumor invades skin and mamilla,
\item Invasiveness of the tumor invades wall or muscle,
\item Number of involved lymph nodes 0,
\item Number of involved lymph nodes between 1 and 3,
\item Number of involved lymph nodes between 4 and 9,
\item Number of involved lymph nodes 10 or more,
\item application of a therapy (cTherapy) \textit{false},
\item application of a therapy (cTherapy) \textit{true},
\item application of a therapy (hTherapy) \textit{false},
\item application of a therapy (hTherapy) \textit{true},
\item Medical history no cancer,
\item Medical history 1st generation breast, ovarian or prostate cancer,
\item Medical history 2nd generation breast, ovarian or prostate cancer,
\item Medical history unknown gynecological cancer,
\item Medical history colon or pancreas cancer,
\item Medical history other or unknown cancers,
\item Medical history not determined,
\item lymphatic or vascular invasion \textit{false}
\item lymphatic or vascular invasion \textit{true}
\item Level of estrogen receptors in tumor (in fmol per mg of protein) less than 5,
\item Level of estrogen receptors in tumor (in fmol per mg of protein) 5 to 10,
\item Level of estrogen receptors in tumor (in fmol per mg of protein) 10 to 30,
\item Level of estrogen receptors in tumor (in fmol per mg of protein) more than 30,
\item Level of estrogen receptors in tumor (in fmol per mg of protein) not determined,
\item Diameter of the largest removed lymph node less than 15 mm,
\item Diameter of the largest removed lymph node between 15 and 20 mm,
\item Diameter of the largest removed lymph node more than 20 mm,
\item Ratio between involved and total lymph nodes removed 0,
\item Ratio between involved and total lymph nodes removed less that $10\%$,
\item Ratio between involved and total lymph nodes removed between 10 and $30\%$,
\item Ratio between involved and total lymph nodes removed over $30\%$,
\item Patient age group under 40,
\item Patient age group 40-50,
\item Patient age group 50-60,
\item Patient age group 60-70,
\item Patient age group over 70 years.
\end{enumerate}
\scriptsize}
\end{document}